\documentclass[11pt,table]{article}
\usepackage[final]{acl}
\usepackage{times}
\usepackage{latexsym}
\usepackage[T1]{fontenc}
\usepackage[utf8]{inputenc}
\usepackage{microtype}
\usepackage{inconsolata}
\usepackage{graphicx}

\usepackage{xcolor}
\usepackage{algorithm}
\usepackage{algpseudocode}
\usepackage{amsmath}
\usepackage{booktabs}
\usepackage{multirow}
\usepackage{amssymb}

\usepackage{xcolor}
\usepackage{tcolorbox}
\usepackage{booktabs}
\usepackage{enumitem}
\usepackage{geometry} 
\tcbuselibrary{skins, breakable}

\title{ACR: Adaptive Context Refactoring via Context Refactoring Operators for Multi-Turn Dialogue}

\author{
\textbf{Jiawei Shen}\textsuperscript{1}, 
\textbf{Jia Zhu}\textsuperscript{1}, 
\textbf{Hanghui Guo}\textsuperscript{1}, 
\textbf{Weijie Shi}\textsuperscript{3},
\textbf{Yue Cui}\textsuperscript{2},
\textbf{Qingyu Niu}\textsuperscript{1},\\
\textbf{Guoqing Ma}\textsuperscript{1}, 
\textbf{Yidan Liang}\textsuperscript{1}, 
\textbf{Jingjiang Liu}\textsuperscript{1},
\textbf{Yilin Wang}\textsuperscript{1},
\textbf{Shimin Di}\textsuperscript{4}, 
\textbf{Jiajie Xu}\textsuperscript{5}\\[0.5em] 
\textsuperscript{1}Zhejiang Normal University, Zhejiang, China \quad \textsuperscript{2}Alibaba Group\\
\textsuperscript{3}Hong Kong University of Science and Technology, Hong Kong, China\\
\textsuperscript{4}Southeast University, Jiangsu, China \quad \textsuperscript{5}Soochow University, Jiangsu, China
}

\begin{document}
\maketitle
\begin{abstract}
Large Language Models (LLMs) have shown remarkable performance in multi-turn dialogue. However, in multi-turn dialogue, models still struggle to stay aligned with what has been established earlier, follow dependencies across many turns, and avoid drifting into incorrect facts as the interaction grows longer. Existing approaches primarily focus on extending the context window, introducing external memory, or applying context compression, yet these methods still face limitations such as \textbf{contextual inertia} and \textbf{state drift}. To address these challenges, we propose the \textbf{A}daptive \textbf{C}ontext \textbf{R}efactoring \textbf{(ACR)} Framework, which dynamically monitors and reshapes the interaction history to mitigate contextual inertia and state drift actively. ACR is built on a library of context refactoring operators and a teacher-guided self-evolving training paradigm that learns when to intervene and how to refactor, thereby decoupling context management from the reasoning process. Extensive experiments on multi-turn dialogue demonstrate that our method significantly outperforms existing baselines while reducing token consumption.
\end{abstract}

\section{Introduction}
Large Language Models (LLMs) have demonstrated remarkable capabilities in language understanding and generation within multi-turn dialogue scenarios \cite{ouyang2022traininglm}. However, as interaction turns increase, maintaining contextual consistency, modeling long-range dependencies, and ensuring factual faithfulness remain prohibitive challenges \cite{dzirietal2022origin,liu2024lost}.

\begin{figure}[ht]
  \centering
  \includegraphics[width=0.95\columnwidth]{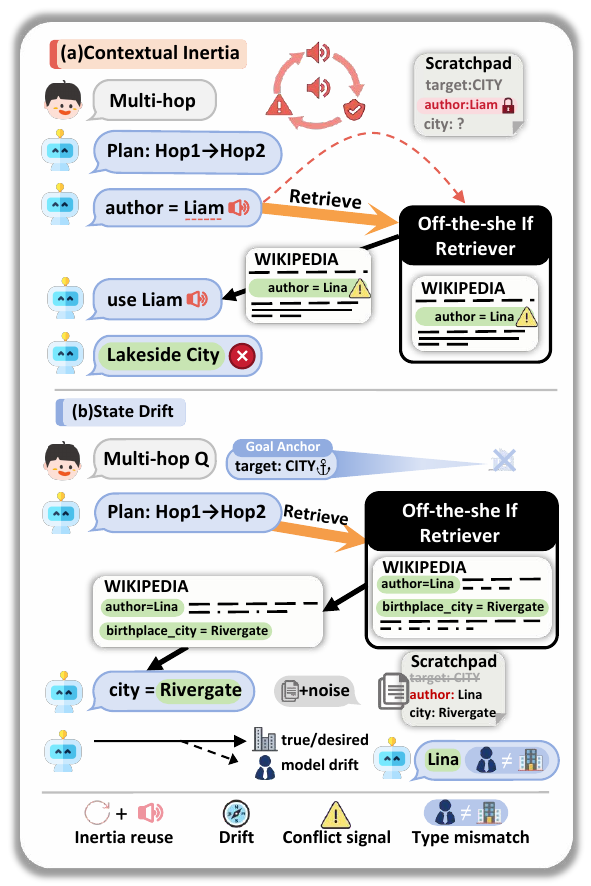}
  \caption{Illustration of two Challenges in multi-turn dialogue: \textit{(a) contextual inertia} and \textit{(b) state drift}.}
  \label{challenge}
\end{figure}
To address these challenges, early studies attempted to enhance the model's ability to handle long-range dialogues by expanding the context window \cite{longlora,flashattention,Cuconasu_2024}. These approaches allow the model to focus on more complete conversation histories, thereby partially alleviating the coherence issues caused by information loss. However, such methods primarily increase the visible information capacity without addressing redundancy in the context. This redundant or even erroneous historical content may be included, potentially introducing noise that hinders subsequent reasoning.

In this context, existing studies attempt to fundamentally address the challenges in multi-turn dialogue by introducing external information and applying context compression. The former seeks to extract critical details from implicit contextual dependencies into explicit, controllable forms. For instance, external memory mechanisms structure and store key information, enabling the model to quickly access and update these details during multi-turn dialogue without relying on lengthy historical context. This helps to avoid information loss or interference caused by excessively long context, thereby maintaining coherence and accuracy in multi-turn dialogue \cite{bae2022keepMemory,Zhong2024MemoryBank}. Meanwhile, Retrieval-Augmented Generation (RAG) queries external knowledge sources to supply explicit evidence, reducing the hallucinations and inconsistencies often caused by volatile memory or missing facts \cite{Lewis2020rag,jiang2022retrieval,su2024dragin,guo-etal-2025-dior}. Conversely, context compression strategies reduce lengthy interactions into shorter, information-dense representations, improving the model's accessibility to key information in subsequent reasoning steps \cite{jiang2023llmlingua,chuang2024learning}. Nevertheless, as shown in Figure~\ref{challenge}, current methods still face challenges:

\textbf{Challenge 1: Contextual Inertia.} Although approaches such as context window extension, external memory, and RAG enrich history and evidence, these methods primarily focus on information provision. They lack adequate coupling with reasoning error-correction mechanisms during generation. In multi-turn dialogue, the model tends to develop path dependency on the existing context, where minor logical deviations or factual errors in earlier turns are continuously incorporated into subsequent reasoning. Since the model is inclined to maintain narrative coherence, errors are not only hard to expose in time but may also be gradually reinforced. Consequently, the model may fall into a locally consistent but globally erroneous cognitive loop, weakening its ability to trace key evidence and revise assumptions.

\textbf{Challenge 2: State Drift.} Existing methods typically represent historical dialogues as unstructured flat sequences. Even with external memory, retrieval, or compression techniques, they still rely on fixed storage and recall strategies, lacking explicit anchoring mechanisms for state evolution. As turns accumulate, initial global constraints and intermediate goals may become diluted by local questions and noise. While these pieces of information may not be significant at the moment, they may become critical at later reasoning nodes. Once this key intermediate information is submerged by noise or lost during compression, subsequent reasoning proceeds based on incomplete state representations. This causes the model to drift from updated constraints, ultimately resulting in logical breaks and failed objectives.

To address the above challenges, we propose an \textbf{A}daptive \textbf{C}ontext \textbf{R}efactoring \textbf{(ACR)} framework, which proactively manages and refactors the interaction history when needed to improve the stability and reliability of multi-turn dialogue. Specifically, to tackle contextual inertia and state drift, we first construct a library of context refactoring operators covering six strategy types. Building on this library, we introduce a \textbf{teacher-guided self-evolving} training paradigm that enables the model to learn when to refactor and how to select and execute refactoring strategies. This paradigm iteratively optimizes the router and the refactorer in a closed loop, resulting in an LLM with monitoring capabilities that continuously diagnose the evolving history context. When signs of drift or inertia are detected, the router selects an appropriate operator to trigger intervention. The refactorer then applies the corresponding strategy to produce a refactored context, which replaces the original history and is fed into the reasoning model for subsequent inference.

Our contributions are summarized as follows:

\begin{itemize}[leftmargin=*]
    \item We propose ACR, which dynamically monitors and reshapes the interaction history to actively mitigate the challenges of contextual inertia and state drift in multi-turn dialogue.
    \item We introduce a library of \textbf{context refactoring operators} and a \textbf{Teacher-Guided Self-Evolving} training paradigm. This method decouples context management from reasoning, enabling the LLM to internalize refactoring capabilities without expensive reinforcement learning.
    \item Extensive experiments on long-context tasks demonstrate that our method significantly outperforms existing baselines while reducing token consumption. The results validate that context refactoring is a more efficient path to long-horizon reasoning than enhancing logic via RL.
\end{itemize}

\begin{figure*}[t]
  \centering
  \includegraphics[width=\linewidth]{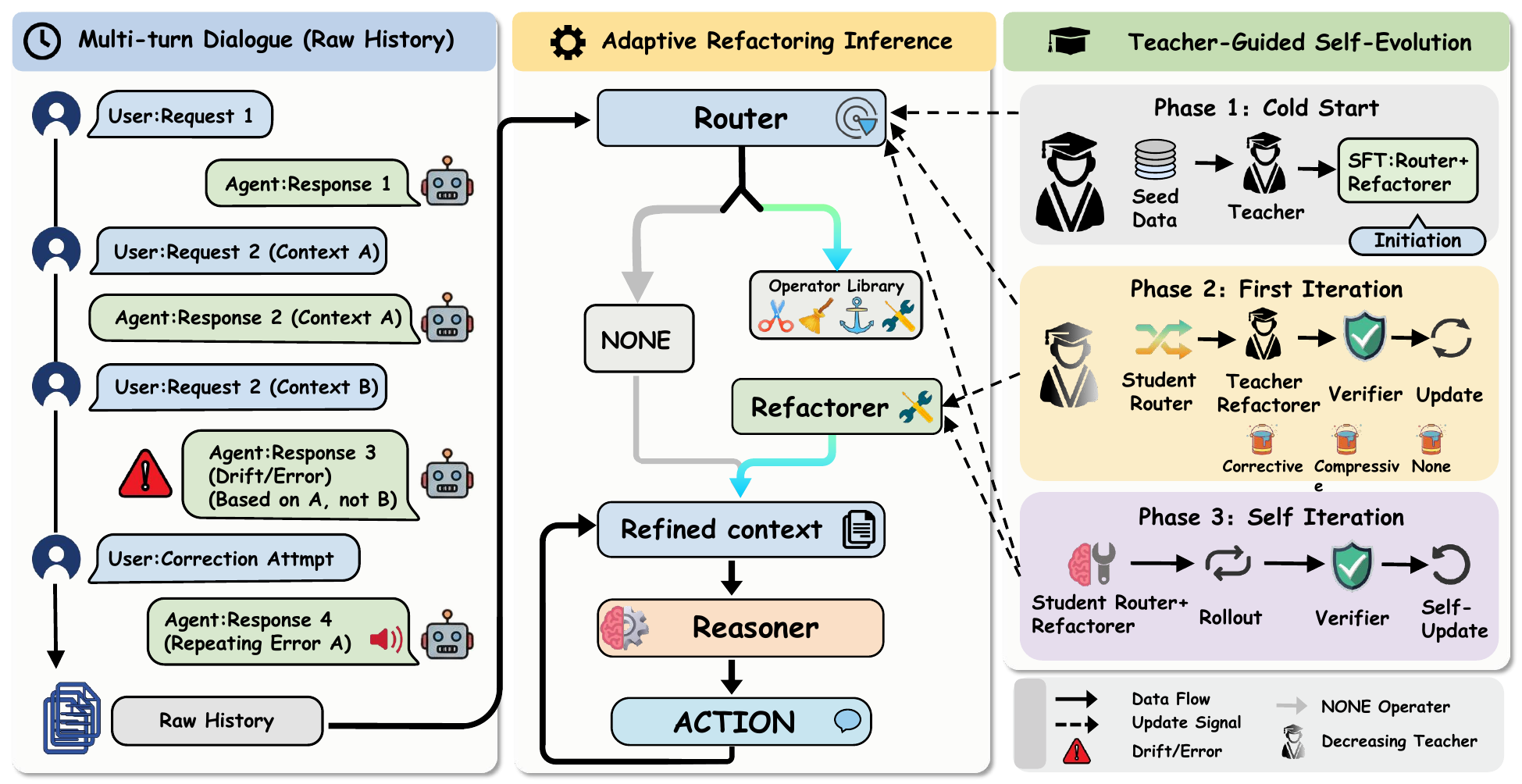}
  \caption{The framework of ACR. The pipeline includes Semantic Routing, where the Router selects the appropriate context operator, and Adaptive Refactoring, which refines the context for improved reasoning. The framework evolves through a Teacher-Guided Self-Evolution training, transitioning from supervised learning to autonomous decision-making, enhancing reasoning accuracy and efficiency.}
  \label{fig:framework}
\end{figure*}

\section{Related Work}

Large language models (LLMs) have exhibited remarkable capabilities in multi-turn dialogue \cite{liu2024lost,bai-etal-2024-longbench}. However, their performance often degrades as the number of interaction turns increases, primarily due to difficulties in maintaining contextual consistency, capturing long-range dependencies, and ensuring factual accuracy.

To overcome these issues, researchers have proposed a variety of strategies. In early studies, methods like DRAGIN~\cite{su2024dragin}, DioR~\cite{guo-etal-2025-dior}, and SEAKR~\cite{yao2025seakr} dynamically trigger retrieval based on uncertainty metrics or classifier-based detection. While effective at supplying explicit evidence, these approaches suffer from \textbf{Contextual Inertia}. They primarily focus on information provision rather than reasoning correction; if the historical context already contains logical deviations, retrieving additional evidence based on flawed premises often reinforces rather than resolves the error loop.

To handle the problem of long horizons, HippoRAG~\cite{jimenez2024hipporag} and ITER-RETGEN~\cite{shao2023itergen} utilize knowledge graphs or iterative retrieval, while RECOMP~\cite{xu2024recomp} employs selective compression to save tokens. However, these methods typically treat history as a static accumulation or reduce it via indiscriminate summarization. Lacking explicit anchoring mechanisms, they are vulnerable to \textbf{State Drift}, where critical global constraints are submerged by local noise or lost during compression, leading to reasoning on incomplete state representations.

Reinforcement Learning methods optimize agent policies for multi-step reasoning. Search-R1~\cite{jin2025search} treats search as an environment with token masking, and StepSearch~\cite{zheng2025stepsearch} achieves process-level supervision for LLM search via step-wise Proximal Policy Optimization incorporating information gain and redundancy penalties, thereby significantly improving reasoning performance on complex multi-hop QA tasks. They lack the ability to actively refactor the context, making them insufficient to break the path dependency inherent in long-context reasoning.

\section{Preliminary} 
\label{sec:formulation}



\noindent \textbf{Multi-turn dialogue.} Formally, we define a multi-turn dialogue session as a sequential process $\mathcal{S} = \{(u_1, a_1), \dots, (u_T, a_T)\}$, where $u_t$ denotes the user instruction at turn $t$, and $a_t$ represents the corresponding agent response. At any time step $t$, the History Context available to the model consists of the accumulated sequence of previous interactions:
\begin{equation}
H_t = [u_1, a_1, \dots, u_{t-1}, a_{t-1}].
\end{equation}

\noindent \textbf{Standard Reasoning Paradigm.} Given an LLM parameterized by $\theta$, the standard objective is to generate the response $a_t$ by maximizing the conditional likelihood given the raw history and the current instruction:
\begin{equation}
a_t = \underset{{a \in \mathcal{V}^*}}{\operatorname{arg max}} P_\theta(a \mid H_t, u_t).
\end{equation}
In this paradigm, the model relies solely on the implicit attention mechanism to extract relevant features from the potentially lengthy and noisy $H_t$.

\noindent \textbf{Objective.} Our goal is to transcend the limitations of conditioning on raw history. We aim to identify an optimized context representation $\tilde{H}_t$ derived from $H_t$, such that the likelihood of generating the optimal response $a_t^*$ is maximized by $\max P_\theta(a_t^* \mid \tilde{H}_t, u_t)$.

In the following sections, we introduce a mechanism to explicitly construct $\tilde{H}_t$ via a library of context refactoring operators.


\section{Proposed Framework: ACR}

In this section, we present \textbf{A}daptive \textbf{C}ontext \textbf{R}efactoring \textbf{(ACR)}, an innovative multi-turn dialogue framework, as illustrated in Figure~\ref{fig:framework}. ACR monitors the evolving history context during inference and refactors it on demand to improve the stability and reliability of multi-turn dialogue. 

\subsection{Adaptive Context Refactoring Inference}
\label{sec:pipeline}




Adaptive context refactoring inference is a core component of the ACR framework, which introduces an external controller model to supervise and refactor the dialogue history on demand. Unlike prior methods that passively concatenate and accumulate dialogue history, ACR preserves the full history $H$ while introducing an external refactoring controller to dynamically supervise the reasoning process. When signals are detected, the controller refactors the history in a \emph{need-driven} and \emph{structured} manner, providing the downstream reasoner with a higher signal-to-noise input. ACR manages the reasoning context via two stages:

\textbf{Semantic Routing.} We instantiate a semantic router that continuously encodes the current history $H$ and predicts a probability distribution over refactoring operators defined in Section~\ref{section:op}.
The router identifies the contextual status of the current turn: whether the history should be summarized, corrected, or left unchanged by selecting the \textsc{none} operator. This stage shifts context handling from indiscriminate full-history feeding to problem-aware strategy selection.

\textbf{Dynamic Refactoring.} When the router selects a non-\textsc{none} operator, the system \emph{hot-swaps} the corresponding Refactoring LoRA,
which transforms the raw history $H$ into a structured and low-noise refactored context $\tilde{H}$ according to the chosen operator.
We then replace the original history with $\tilde{H}$ as the single, coherent information source for the downstream reasoning model.

This closed-loop design reduces the attention and retrieval burden over long histories and suppresses the repeated reuse and rationalization of early mistakes, thereby improving the stability and controllability of multi-turn dialogue.

\subsection{The Library of Context Refactoring Operators}
\label{section:op}
In multi-turn dialogue, the raw interaction history $H$ is often laden with redundancy, stochastic noise, hallucinations, or logical deadlocks. Directly conditioning an LLM on such raw $H$ exacerbates state drift. To address this, we define a comprehensive library of \textbf{Context Refactoring Operators}, denoted as $\mathcal{A}$. These operators are not merely heuristic truncation rules but are grounded in information theory, aiming to transform linear, flat logs into structured, high-semantic-density memory representations. We categorize the six operators into three distinct functional groups: Information Density Optimization, Logical Flow Control, and Attention Management. Additionally, an \textbf{identity operator} is included to handle optimal contexts.

\subsubsection{Category 1: Information Density Optimization}

\noindent \textbf{State Abstraction.} In multi-turn dialogue, intermediate steps often contain redundant exploration that no longer influences future decisions. Relying on the Markov Assumption, $\mathcal{O}_{\text{abs}}$ compresses the serialized action-observation history into a semantic State Snapshot, while retaining the most recent user instruction to ensure task continuity.

\begin{equation}
\mathcal{O}_{\text{abs}}(H) = \mathcal{S}_{\phi}(H_{<t}) \oplus x_t,
\end{equation}
where $\mathcal{S}_{\phi}$ denotes the state summarization function, $x_t$ is the current query, and $\oplus$ represents concatenation.

\noindent \textbf{Noise Filtering.} Retrieved contexts often contain distractor documents that are orthogonal to the current intent. $\mathcal{O}_{\text{filter}}$ performs surgical text pruning by evaluating the semantic relevance of each text unit $u_i$ within $H$, filtering out segments that fall below a relevance threshold $\tau$:

\begin{equation}
    \mathcal{O}_{\text{filter}}(H, x_t) = u_i,
\end{equation}
where $u_i \in \{H \mid \text{Sim}(u_i, x_t) > \tau\}$. This operator maximizes the signal-to-noise ratio within the limited context window.

\subsubsection{Category 2: Logical Flow Control}

\noindent\textbf{Fact Rectification.} Long-context generation is prone to hallucinations and the persistence of obsolete or incorrect claims. Unlike standard approaches that append corrections to the end, $\mathcal{O}_{\text{rect}}$ utilizes an external verifier $\mathcal{V}$ to identify hallucinated propositions and performs in-place editing.
Specifically, an external verifier $\mathcal{V}$ checks each context unit $u_i$ and decides whether it is factually valid.
If $u_i$ is verified as true, it is kept unchanged; otherwise, it is rewritten by a rewriting function $\mathcal{R}$.

\begin{equation}
\begin{split}
    \mathcal{O}_{\text{rect}}(H) = [\tilde{u}_1, \dots, \tilde{u}_n], \\ \text{where } H=[u_1,\dots,u_n],
\end{split}
\end{equation}

\begin{equation}
\tilde{u}_i =
\begin{cases}
u_i, & \text{if } \mathcal{V}(u_i)=\texttt{true},\\
\mathcal{R}(u_i), & \text{otherwise}.
\end{cases}
\end{equation}

\noindent \textbf{Path Pruning.} Complex reasoning often leads to logical loops or incorrect exploration branches. Analogous to backtracking in search algorithms, $\mathcal{O}_{\text{prune}}$ identifies the point of logical bifurcation or failure, denoted as $k$. It explicitly truncates the history after $k$, rolling the context back to the nearest clean state to prevent error cascading:

\begin{equation}
\mathcal{O}_{\text{prune}}(H_{1:t}) = H_{1:k},
\end{equation}
where $k < t$ is the divergence index.

\subsubsection{Category 3: Attention Management}

\noindent \textbf{Cognitive Boosting.} Even with complete information, models may face a reasoning gap in translating context into action. $\mathcal{O}_{\text{boost}}$ injects a Chain-of-Thought or sub-goal definition at the end of the context. This acts as a system hint, explicitly bridging the implicit logic required for the next step:

\begin{equation}
\begin{split}
\mathcal{O}_{\text{boost}}(H) = H \oplus z_{\text{thought}}, \\ z_{\text{thought}} \sim P_{\text{CoT}}(\cdot|H).
\end{split}
\end{equation}

\noindent \textbf{Key Anchoring.} Addressing the Lost-in-the-Middle phenomenon, $\mathcal{O}_{\text{anchor}}$ exploits the Recency Bias of LLMs. It identifies global constraints or critical instructions $c_{key}$ that may have been diluted by subsequent turns and copies them to the Active Attention Zone:

\begin{equation}
\mathcal{O}_{\text{anchor}}(H) = H \oplus \text{``[REMINDER]: ''} \oplus c_{key}.
\end{equation}

\subsubsection{The Identity Operator}

Finally, we define $\mathcal{O}_{\text{NONE}}(H) = H$, which is selected when the Router determines that the current context requires no intervention.

\subsection{Teacher-Guided Self-Evolving Training Paradigm}

Training the model to effectively balance operator selection and context rewriting is non-trivial. To address this, we propose \textbf{Teacher-Guided Self-Evolving (TGSE)} Training Paradigm, a progressive training framework that transitions the model from supervised imitation to autonomous self-evolution. The training pipeline consists of three distinct phases:

\paragraph{Phase I: Supervised Initialization.} To mitigate the instability of random exploration in the early stages, we cold-start both the Router policy $\pi_\theta$ and the Refactorer $\phi_\omega$ with a small seed set $\mathcal{D}_{\text{seed}}$. Specifically, we employ a strong teacher model to generate high-quality supervision for routing decisions and corresponding refactored contexts $\hat{H}_t$. We then perform supervised fine-tuning on these teacher-labeled instances, yielding an initial router that can reliably identify when refactoring is needed and an initial refactorer that can execute basic context edits with high fidelity.

\paragraph{Phase II: Teacher-Guided Trajectory Rollout.} We bootstrap high-quality corrective supervision with a teacher-in-the-loop rollout procedure. Given the current history $H_t$, the student Router $\pi_\theta$ first samples an intervention decision. To avoid propagating errors from an immature Refactorer, we then delegate the execution of the selected operator to a strong teacher model, producing a higher-fidelity refactored context $\hat{H}_{\text{teacher}}$. We finally perform hindsight verification by running the base solver on both $H_t$ and $\hat{H}_{\text{teacher}}$ and measuring the resulting task outcome. We keep only the cases where teacher refactoring yields a clear improvement and add them as positive training instances for subsequent self-evolution.

\paragraph{Phase III: Autonomous Evolution.} As the local Refactoring module $\phi_\omega$ matures, we progressively decouple the system from the teacher. The framework enters a Bootstrapping mode, where the system samples and verifies trajectories using locally generated contexts $\hat{H}_{local}$. This facilitates the internalization of refactoring capabilities and enables closed-loop iteration.

\paragraph{Dynamic Data Synthesis Strategy.} To ensure a balanced learning objective, the training data for self-evolution is dynamically composed of three distinct categories:

\textbf{Corrective Instances.} We form corrective pairs where the model fails under the raw history but succeeds after refactoring:
\begin{equation}
    (H_t, y_{\text{fail}})\ \rightarrow\ (\hat{H}_t, y_{\text{success}}).
\end{equation}
These samples provide the most informative supervision for both modules: they encourage the Router to trigger refactoring when the current context exhibits drift or accumulated noise, and train the Refactorer to remove misleading or stale information that causes failure. Accordingly, we assign them a higher loss weight to emphasize failure-to-success transitions.

\textbf{Compressive Instances.} We create compressive pairs that preserve task success while reducing context length:
\begin{equation}
\begin{split}
    (H_t, y_{\text{success}})\ \rightarrow\ (\hat{H}_t, y_{\text{success}}), \\ \text{s.t. } |\hat{H}_t| \ll |H_t|.
\end{split}
\end{equation}
These instances teach the model to retain only the information necessary for correct reasoning, improving efficiency via higher information density without degrading accuracy.

\textbf{Regularization Instances.} For contexts that are already clean and unambiguous, we include non-intervention examples:
\begin{equation}
    (H_t, y_{\text{success}})\ \rightarrow\ (\text{Action: None}).
\end{equation}
They explicitly discourage unnecessary edits, preventing an over-refactoring tendency and stabilizing performance on easy or low-noise cases.


\begin{algorithm}[t]
\caption{TGSE Training}
\label{alg:tgse_en}
\begin{algorithmic}[1]
\State \textbf{Input:} Seed set $\mathcal{D}_{\text{seed}}$, operators $\mathcal{O}\cup\{\textsc{None}\}$, feedback $R(\cdot)$
\State \textbf{Phase I (Cold Start):} Use a teacher to generate supervision on $\mathcal{D}_{\text{seed}}$; SFT Router $\pi_\theta$ and Refactorer $\phi_\omega$.
\State \textbf{Phase II+ (Self-Evolution):}
\For{each iteration}
    \State sample a task segment and obtain $H_t$
    \State sample $o_t\sim \pi_\theta(\cdot\mid H_t)$
    \State obtain $\hat{H}\leftarrow \textsc{Teacher}(H_t,o_t)$ with prob. $p_{\text{teacher}}$, else $\hat{H}\leftarrow \phi_\omega(H_t,o_t)$
    \State compute $R(H_t)$ and $R(\hat{H})$
    \If{$R(\hat{H})\ge R(H_t)+\delta$}
        \State push into $\mathcal{B}_{\text{corr}}$
    \ElsIf{$R(\hat{H})\approx R(H_t)\ \land\ |\hat{H}|\ll |H_t|$}
        \State push into $\mathcal{B}_{\text{comp}}$
    \Else
        \State push into $\mathcal{B}_{\text{reg}}$ (supervise $\textsc{None}$)
    \EndIf
    \State sample minibatch from pools by fixed ratios and update $\pi_\theta,\phi_\omega$
    \State anneal $p_{\text{teacher}}\downarrow$
\EndFor
\end{algorithmic}
\end{algorithm}

\section{Experiments}
In this section, we first introduce our experimental setup, and then report the main results, ablation studies, and efficiency analysis to comprehensively validate the effectiveness of the proposed \textsc{ACR} framework. More details of the experiments can be seen in Appendix~\ref{ap:Experiments}.

\begin{table*}[htbp]
  \centering
  \caption{Main results (EM, \%) on seven QA benchmarks (single-hop and multi-hop). We compare ACR with baselines under the same retriever and backbone. ${\dagger}$ denotes in-domain datasets and ${\star}$ denotes out-of-domain datasets.}
  \label{tab:qa_results}
  \renewcommand{\arraystretch}{1.1}

  \newcommand{\best}[1]{\textbf{#1}}
  \newcommand{\second}[1]{\underline{#1}}

  \resizebox{\textwidth}{!}{%
    \begin{tabular}{l|l|ccccccc}
    \toprule
    \multirow{2}{*}{\textbf{Type}} & \multirow{2}{*}{\textbf{Method}} &
    \multicolumn{3}{c}{\textbf{Single-Hop QA}} & \multicolumn{4}{c}{\textbf{Multi-Hop QA}} \\
    \cmidrule(lr){3-5} \cmidrule(lr){6-9}
     & & \multicolumn{1}{c}{NQ$^{\dagger}$} & \multicolumn{1}{c}{TriviaQA$^{\star}$} & \multicolumn{1}{c}{PopQA$^{\star}$}
       & \multicolumn{1}{c}{HotpotQA$^{\dagger}$} & \multicolumn{1}{c}{2Wiki$^{\star}$} & \multicolumn{1}{c}{MuSiQue$^{\star}$} & \multicolumn{1}{c}{Bamboogle$^{\star}$} \\
    \midrule
    \multirow{3}{*}{Prompt} & Direct inference & 13.40 & 40.80 & 14.00 & 18.30 & 25.00 & 3.10  & 12.00 \\
          & CoT   & 4.80  & 18.50 & 5.40  & 9.20  & 11.10 & 2.20  & 23.20 \\
          & IRCoT & 22.40 & 47.80 & 30.10 & 13.30 & 14.90 & 7.20  & 22.40 \\
    \midrule
    SFT   & SFT   & 31.80 & 35.40 & 12.10 & 21.70 & 25.90 & 6.60  & 11.20 \\
    \midrule
    \multirow{3}{*}{RAG} & DRAGIN & 23.20 & 42.00 & \multicolumn{1}{c}{--} & 23.20 & 22.00 & \multicolumn{1}{c}{--} & \multicolumn{1}{c}{--} \\
          & DioR  & 26.20 & 52.30 & \multicolumn{1}{c}{--} & 27.40 & 26.60 & \multicolumn{1}{c}{--} & \multicolumn{1}{c}{--} \\
          & SEAKR & 25.60 & 54.40 & \multicolumn{1}{c}{--} & 27.90 & 30.20 & \multicolumn{1}{c}{--} & \multicolumn{1}{c}{--} \\
    \midrule
    Compression & RECOMP (Flan-UL2-20B) & \second{36.60} & \second{58.99} & \multicolumn{1}{c}{--} & 30.40 & \multicolumn{1}{c}{--} & \multicolumn{1}{c}{--} & \multicolumn{1}{c}{--} \\
    \midrule
    \multirow{2}{*}{Mem.} & HippoRAG (GPT-3.5) & \multicolumn{1}{c}{--} & \multicolumn{1}{c}{--} & \multicolumn{1}{c}{--} & \best{45.70} & \best{47.70} & 21.90 & \multicolumn{1}{c}{--} \\
          & ITER-RETGEN & \multicolumn{1}{c}{--} & \multicolumn{1}{c}{--} & \multicolumn{1}{c}{--} & \second{45.20} & 35.50 & \best{25.90} & \best{40.00} \\
    \midrule
    \multirow{3}{*}{RL Training} & R1-Instruction & 21.00 & 44.90 & 17.10 & 20.80 & 27.50 & 6.00  & 19.20 \\
          & Search-R1 & \best{39.30} & \best{61.00} & \best{39.70} & 37.00 & \second{40.10} & 14.60 & \second{36.80} \\
          & StepSearch & -- & -- & -- & 38.60 & 36.60 & \second{22.60} & \best{40.00} \\
    \midrule
    \rowcolor{gray!20}
    SFT & \textbf{ACR (Ours)} & 36.41 & 56.86 & \second{36.04} & 35.10 & 34.32 & 16.67 & 36.36 \\
    \bottomrule
    \end{tabular}%
  }
\end{table*}

\subsection{Experimental Setups}
\label{sec:es}
\textbf{Datasets.} We evaluate our framework on \emph{multi-turn QA} benchmarks that stress long-range dependency tracking and evidence aggregation. Following prior agent-style QA settings, we build the training set by merging \textbf{Natural Questions (NQ)} (single-hop) and \textbf{HotpotQA} (multi-hop) \cite{jin2025search,zheng2025stepsearch}, and assess generalization on a suite of \textbf{seven} QA datasets: single-hop QA (\textbf{NQ}~\cite{kwiatkowski2019natural}, \textbf{TriviaQA}~\cite{joshi2017triviaqa}, and \textbf{PopQA}~\cite{mallen2023not}) and multi-hop QA (\textbf{HotpotQA}~\cite{yang2018hotpotqa}, \textbf{2Wiki}~\cite{ho2020constructing}, \textbf{MusiQue}~\cite{trivedi2022musique}, and \textbf{Bamboogle}~\cite{press2023measuring}). See Appendix~\ref{app:data} for the full list and dataset statistics. We use \textbf{Exact Match (EM)} as the metric.

\textbf{Baselines.} We compare our method against representative baselines across several paradigms, including \textbf{prompting} (\textsc{Direct inference}, \textsc{CoT} \cite{wei2022chain}, \textsc{IRCoT} \cite{trivedi2023interleaving}), \textbf{SFT} \cite{chung2024scaling}, \textbf{retrieval-augmented QA} (\textsc{DRAGIN} \cite{su2024dragin}, \textsc{DioR} \cite{guo-etal-2025-dior}, \textsc{SEAKR} \cite{yao2025seakr}), \textbf{context compression} (\textsc{RECOMP} \cite{xu2024recomp}), \textbf{external memory} (\textsc{HippoRAG} \cite{jimenez2024hipporag}, \textsc{ITER-RETGEN} \cite{shao2023itergen}), and \textbf{RL-style search training} (\textsc{R1-Instruction} \cite{guo2025deepseek}, \textsc{Search-R1} \cite{jin2025search}, \textsc{StepSearch} \cite{zheng2025stepsearch}); full baseline details are deferred to Appendix~\ref{app:baseline}. 

\textbf{Train Details.} To ensure a controlled comparison, we adopt the same retriever setting as Search-R1 and use E5 as the retriever. We use \textbf{Qwen-2.5-7B-Instruct} as the downstream reasoner, GPT-5.2 as teacher model and instantiate both the \textbf{Router} and \textbf{Refactorer} on the same backbone for parameter sharing and fair capacity matching. We train the Router and Refactorer with parameter-efficient adapters under the proposed TGSE training paradigm, and implement all iterative updates using \textbf{LLaMA-Factory}. Hyperparameter settings are reported in Appendix~\ref{app:td}.

\subsection{Experimental Results}
\label{sec:mr}
In this section, we report our main experimental results to validate the effectiveness of \textsc{ACR} across diverse datasets. We further study the contribution of each component and analyze the efficiency of our approach. Additional experimental results and extended analyses are provided in the Appendix~\ref{ap:Experiments}.

\subsubsection{Overall Experiments}
Table~\ref{tab:qa_results} reports EM results of ACR on seven QA benchmarks. We compare ACR with prompting (Direct/CoT/IRCoT), SFT, RAG variants, compression/external-memory methods, and RL-style search training. The results support four observations.
\textbf{(1) Consistent gains over conventional paradigms.} ACR yields stable improvements over prompting and vanilla SFT on all datasets. In particular, it improves SFT from 31.80$\rightarrow$36.41 on NQ, 35.40$\rightarrow$56.86 on TriviaQA, 12.10$\rightarrow$36.04 on PopQA, and 21.70$\rightarrow$35.10 on HotpotQA, suggesting that need-driven history refactoring is broadly effective for both single-hop and multi-hop QA.
\textbf{(2) Interpreting compression/external-memory baselines.} Several compression/memory baselines report strong results on multi-hop settings, but their advantage is partly driven by stronger backbones in their default configurations (e.g., RECOMP with Flan-UL2-20B; HippoRAG with GPT-3.5). 
\textbf{(3) Low-cost competitiveness vs.\ RL-style training.} RL-based search training remains highly competitive. Nevertheless, ACR narrows the gap with substantially lower training overhead: it trains only an external routing/refactoring controller using 3.8K supervised instances, compared with 170K for Search-R1 and 19K for StepSearch, while avoiding online rollouts, reward engineering, and policy optimization. Despite this lightweight setup, ACR stays close to Search-R1 and surpasses it on MuSiQue (16.67 vs.\ 14.60), demonstrating favorable cost--performance trade-offs.
\textbf{(4) Modularity and generality.} Notably, ACR updates only the external controller while keeping the underlying reasoner fixed, yet consistently improves performance across diverse QA benchmarks, highlighting the generality of plug-in context refactoring.

\begin{table*}[htbp]
  \centering
  \caption{Ablation results of \textsc{ACR} (EM, \%). We disable the Router or Refactorer to assess their contributions across QA benchmarks. Where ``Base'' refers to using Qwen2.5-7B-instruct as the router and refactorer.}
  \label{tab:ablation_all}
  \renewcommand{\arraystretch}{1.1}
    \begin{tabular}{l|ccccccc}
    \toprule
    \multirow{2}{*}{\textbf{Variant}} & \multicolumn{3}{c}{\textbf{Single-Hop QA}} & \multicolumn{4}{c}{\textbf{Multi-Hop QA}} \\
    \cmidrule(lr){2-4} \cmidrule(lr){5-8}
          & \multicolumn{1}{c}{NQ$^{\dagger}$} & \multicolumn{1}{c}{TriviaQA$^{\star}$} & \multicolumn{1}{c}{PopQA$^{\star}$} & \multicolumn{1}{c}{HotpotQA$^{\dagger}$} & \multicolumn{1}{c}{2Wiki$^{\star}$} & \multicolumn{1}{c}{MuSiQue$^{\star}$} & \multicolumn{1}{c}{Bamboogle$^{\star}$} \\
    \midrule
    Base & 29.56    & 47.49    & 22.16    & 18.43    & 18.77    & 8.48    & 15.74 \\
    w/o Router & 31.03    & \underline{48.34}    & 25.32    & 21.28    & 20.78    & 9.78    & 23.45 \\
    w/o Refactorer & \underline{34.38}    & 47.62    & \underline{30.43}    & \underline{24.53}    & \underline{23.97}    & \underline{10.65}    & \underline{}27.56 \\
    \midrule
    \rowcolor{gray!20} \textbf{Ours} & \textbf{36.41} & \textbf{56.86} & \textbf{36.04} & \textbf{35.10} & \textbf{34.32} & \textbf{16.67} & \textbf{36.36} \\
    \bottomrule
    \end{tabular}%
\end{table*}%

\begin{figure}[ht]
  \centering
  \includegraphics[width=1\columnwidth]{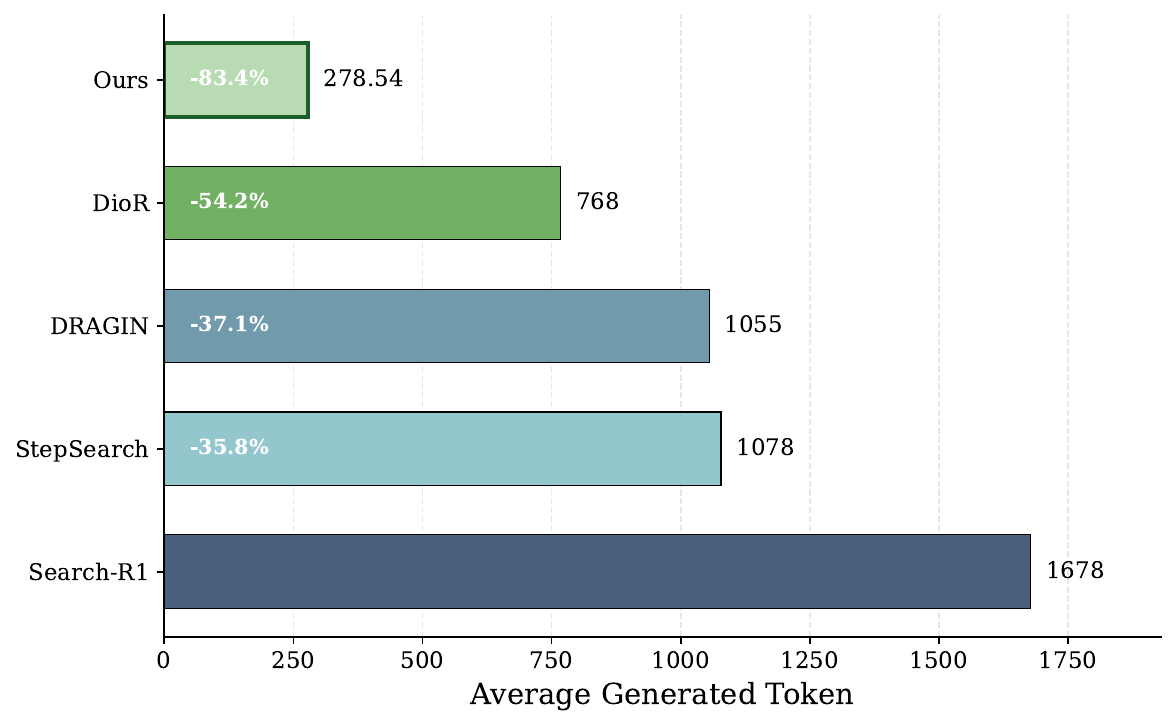}
  \caption{Comparison of average generated tokens across different methods.}
  \label{token}
\end{figure}

\subsubsection{Efficiency Analysis}
We evaluate the efficiency of our approach by analyzing the average number of generated tokens per turn, as shown in Figure~\ref{token}. Compared to Reinforcement Learning (RL) based methods such as \textbf{Search-R1}, our approach demonstrates a significant advantage in computational efficiency. While Search-R1 incurs a high cost averaging \textbf{1678} tokens, our method reduces this consumption to \textbf{278.54} tokens (\textbf{-83.4\%}). Although our method may slightly trail RL approaches in raw performance metrics, the marginal difference is outweighed by this massive gain in efficiency, making it a more practical solution for resource-constrained environments. 

Furthermore, when compared to traditional methods like \textbf{DioR} and \textbf{DRAGIN}, our approach proves to be both \textbf{stronger and more efficient}. We not only achieve lower token consumption but also deliver superior reasoning capabilities, demonstrating that our method effectively eliminates redundant steps without compromising solution quality.

\subsubsection{Ablation Study}
Table~\ref{tab:ablation_all} reports ablations of \textsc{ACR}. We additionally include a \textbf{Base} controller baseline, where both the Router and Refactorer are instantiated by the \textbf{Qwen2.5-7B-Instruct} (prompt-only, \emph{without} training). Disabling either module yields clear degradations across both single-hop and multi-hop benchmarks. The largest drops come from removing the \textbf{Router}: performance decreases by \textbf{13.82} EM on HotpotQA (35.10$\rightarrow$21.28), \textbf{13.54} EM on 2Wiki (34.32$\rightarrow$20.78), and \textbf{12.91} EM on Bamboogle (36.36$\rightarrow$23.45). This indicates that \emph{when-to-intervene} (online drift detection and intervention triggering) is crucial, especially for multi-hop reasoning where early deviations can propagate non-locally. 

The \textbf{Refactorer} is also indispensable: removing it still causes substantial losses (e.g., \textbf{10.57} EM on HotpotQA and \textbf{9.24} EM on TriviaQA), suggesting that accurate detection alone is insufficient without effective \emph{how-to-fix} execution that denoises and restructures the history. Overall, \textsc{ACR} reaches its best performance only when the Router and Refactorer operate in a closed loop, coupling diagnosis with corrective refactoring.

\section{Conclusion}

In this paper, we first investigate the limitations of large language models in multi-turn dialogue as well as prior work that has attempted to address these limitations. However, current methods continue to face significant challenges: \textbf{(1)Contextual Inertia} and \textbf{(2)State Drift}. To overcome these problems, we propose an innovative framework \textbf{A}daptive \textbf{C}ontext \textbf{R}efactoring \textbf{(ACR)}, that actively manages the evolving history instead of passively concatenating it. ACR uses an external controller to monitor the interaction history, select a refactoring operator, and rewrite the context into a cleaner, task-relevant form, thereby decoupling context management from task reasoning. We further introduce a teacher-guided self-evolving training scheme to iteratively improve the router and refactorer. Experiments on single-hop and multi-hop QA benchmarks show consistent gains over strong baselines. In future work, we will internalize refactoring skills into a single unified model.

\section*{Limitations}
Our framework improves multi-turn reasoning by introducing an external controller to monitor and refactor the dialogue history when needed. This design, however, comes with additional deployment overhead. In practice, ACR requires extra modules and thus increases system complexity, memory footprint, and inference latency, even though it can reduce the token budget of the downstream reasoner. As a result, the overall efficiency trade-off may vary across hardware settings and latency constraints. A promising direction is to further internalize refactoring capability into a single model, e.g., by distillation or unified training, so that context refactoring and task reasoning can be performed within one model without relying on an external controller.

In addition, our current evaluation is mainly conducted on QA-style multi-turn reasoning. While these benchmarks capture long-range dependency tracking and factual consistency, they do not fully represent agentic settings that require long-horizon planning, tool use, and interaction with dynamic environments. Therefore, it remains unclear how ACR will behave under more complex agentic workloads, where errors can compound through actions and observations. In future work, we plan to conduct a more comprehensive study on multi-turn agent tasks and interactive environments, and refine the framework to better accommodate task-specific constraints and feedback signals.

\section*{Ethics Statement}
This work utilizes publicly available standard benchmark datasets for evaluation and training, including Natural Questions, HotpotQA, TriviaQA, PopQA, 2WikiMultiHopQA, MuSiQue, and Bamboogle. These datasets are primarily derived from public knowledge sources (e.g., Wikipedia) and do not contain personally identifiable information (PII) or sensitive personal data.

The training process employs a Teacher-Guided Self-Evolving paradigm that relies on synthetic supervision generated by LLMs, without involving new human subject experiments or crowdsourced annotation. Furthermore, the proposed Adaptive Context Refactoring framework aims to enhance the reliability and factual consistency of multi-turn dialogue systems by actively mitigating hallucinations, thereby contributing to the development of safer and more robust AI systems.

\bibliography{custom}

\clearpage

\appendix
This appendix provides supplementary material to ensure the clarity, depth, and reproducibility of our work. It is structured as follows:

\section{Experiments Details}
\label{ap:Experiments}

\subsection{Data Statistics}
\label{app:data}
We construct a training pool by merging two QA sources: Natural Questions (NQ) and HotpotQA, resulting in \textbf{169,615} training instances in total, with \textbf{79,168} (46.7\%) from NQ and \textbf{90,447} (53.3\%) from HotpotQA.
For evaluation, we curate a seven-dataset benchmark suite with \textbf{51,713} examples, covering both single-hop and multi-hop QA:
PopQA \textbf{14,267} (27.6\%), 2WikiMultiHopQA \textbf{12,576} (24.3\%), TriviaQA \textbf{11,313} (21.9\%), HotpotQA \textbf{7,405} (14.3\%), NQ \textbf{3,610} (7.0\%), MuSiQue \textbf{2,417} (4.7\%), and Bamboogle \textbf{125} (0.2\%).
Notably, although the training pool is large, our self-evolving training uses only a small subset: we start with a cold-start set of \textbf{400} examples, and then sample \textbf{200} examples per iteration for \textbf{17} subsequent iterations (18 rounds in total), yielding \textbf{400 + 17$\times$200 = 3,800} training examples overall (about \textbf{2.24\%} of the full training pool).
This design allows us to study the sample efficiency of ACR under a strictly limited supervision budget.

\begin{table}[h]
\centering
\small
\setlength{\tabcolsep}{2pt}
\begin{tabular}{l l r r}
\toprule
Split & Dataset / Source & \#Examples & Share (\%) \\
\midrule
Training & NQ & 79{,}168 & 46.7 \\
Training & HotpotQA & 90{,}447 & 53.3 \\
\midrule
Eval & PopQA & 14{,}267 & 27.6 \\
Eval & 2WikiMultiHopQA & 12{,}576 & 24.3 \\
Eval & TriviaQA & 11{,}313 & 21.9 \\
Eval & HotpotQA & 7{,}405 & 14.3 \\
Eval & NQ & 3{,}610 & 7.0 \\
Eval & MuSiQue & 2{,}417 & 4.7 \\
Eval & Bamboogle & 125 & 0.2 \\
\midrule
\multicolumn{2}{l}{Training budget used in TGSE} & 3{,}800 & 2.24 \\
\bottomrule
\end{tabular}
\caption{Dataset statistics. The training pool is formed by merging NQ and HotpotQA, while evaluation spans seven QA benchmarks.}
\label{tab:data_stats}
\end{table}

\subsection{Baselines}
\label{app:baseline}
For single-hop and multi-hop QA, we compare our approach with a diverse set of competitive baselines:

(1) \textbf{Prompting baselines:} \textsc{Prompt}, \textsc{CoT}, and \textsc{IRCoT}, which rely purely on in-context prompting (with IRCoT further incorporating iterative retrieval into the reasoning trace) without updating model parameters.

(2) \textbf{Supervised fine-tuning:} \textsc{SFT}, a task-adapted baseline trained with standard supervised learning.

(3) \textbf{Retrieval-augmented QA:} \textsc{DRAGIN}, \textsc{DioR}, and \textsc{SEAKR}, which dynamically trigger retrieval and augment generation with external evidence.

(4) \textbf{Context compression:} \textsc{RECOMP}, which improves long-context efficiency by selectively compressing and augmenting contexts.

(5) \textbf{External memory methods:} \textsc{HippoRAG} and \textsc{ITER-RETGEN}, which leverage long-term memory structures / iterative retrieval-generation to better support multi-hop evidence aggregation.

(6) \textbf{RL-style search training:} \textsc{R1-Instruct}, \textsc{Search-R1}, and \textsc{StepSearch}, which optimize search-augmented reasoning policies via reinforcement learning-style training signals.

\subsection{Training Details}
\label{app:td}

\textbf{Backbone models.} The Router and Refactoring modules are implemented as lightweight LoRA adapters attached to an external controller model. For the downstream solver (reasoner), we use \textsc{Qwen-2.5-7B-Instruct} as a frozen backbone. We do not update the solver parameters during training.
Unless otherwise specified, we decode with temperature $0.7$ and set the maximum generation length to $8192$ tokens.

\textbf{Training hyperparameters.} We employ the AdamW optimizer with a learning rate of $2 \times 10^{-4}$ and a per-device batch size of 4. For parameter-efficient fine-tuning, we utilize LoRA with rank $r=16$ and $\alpha=32$. The training process is set to run for 3 epochs per iteration. To prevent overfitting, we apply an early-stopping criterion on the validation set with a patience of 3 iterations and a minimum improvement threshold (min\_delta) of $0.001$. Regarding the TGSE knobs: we set the early stopping delta $\delta=0.001$. Note that the pool ratios and $p_{\text{teacher}}$ annealing schedule are not explicitly defined in the provided configuration and may require manual specification based on the implementation details.

\textbf{Hardware.} All experiments are conducted on 2$\times$ NVIDIA A100 GPUs (80GB each). We use LoRA training to reduce the memory footprint and accelerate training.

\subsection{Training Loss}
\label{app:loss}
\begin{figure}[ht]
  \centering
  \includegraphics[width=0.95\columnwidth]{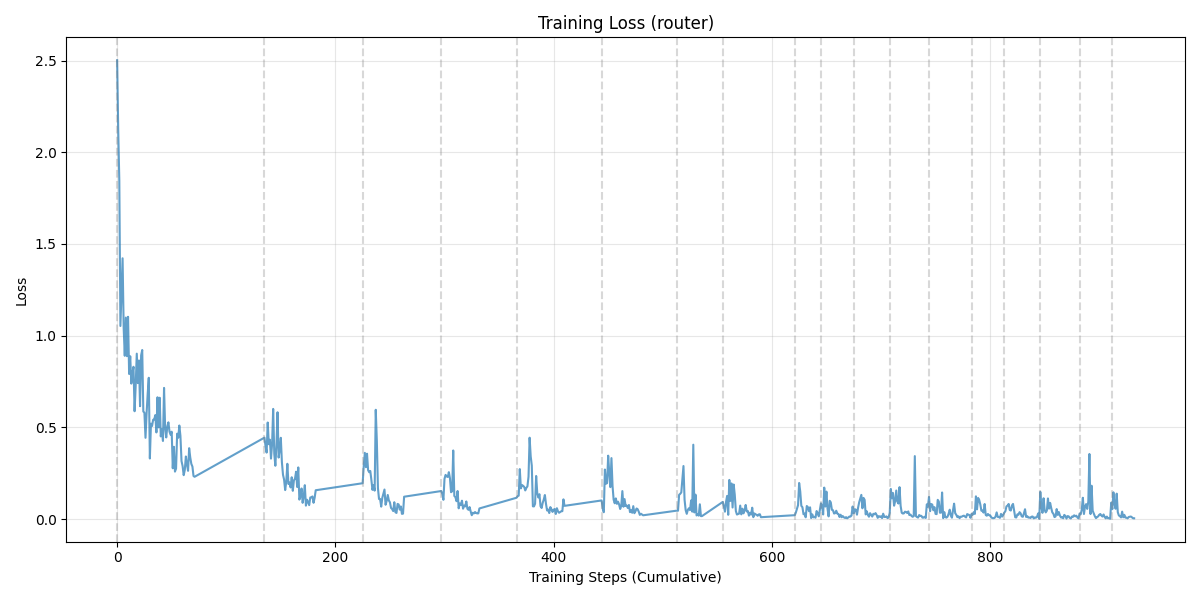}
  \caption{Training loss of the Router (LoRA) across TGSE rounds.}
  \label{router}
\end{figure}

\begin{figure}[ht]
  \centering
  \includegraphics[width=0.95\columnwidth]{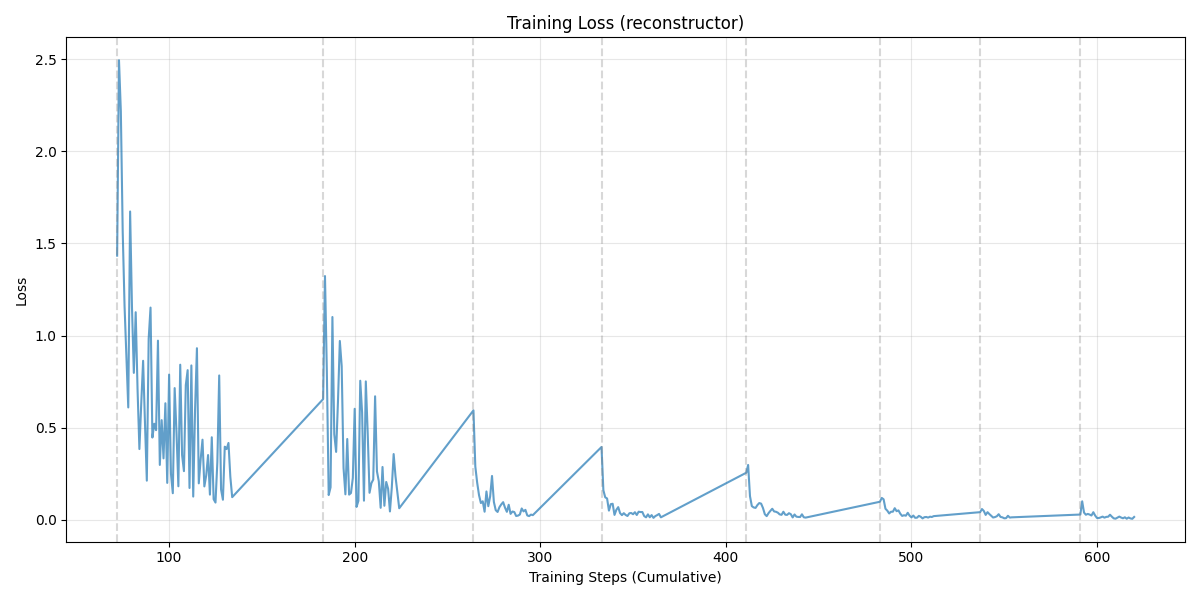}
  \caption{Training loss of the Refactorer (LoRA) across TGSE rounds.}
  \label{refactorer}
\end{figure}
\paragraph{Training dynamics.}
Figure~\ref{router} and Figure~\ref{refactorer} illustrate the training loss trajectories of the Router and Refactorer (LoRA) under TGSE.
Both modules exhibit a clear convergence pattern: the loss drops sharply during the cold-start stage, indicating that a small amount of supervision is sufficient to learn effective routing decisions and basic refactoring behaviors.
Across subsequent self-evolving rounds, the curves show short-lived, iteration-aligned spikes (marked by dashed vertical lines), which we attribute to mild distribution shifts when newly sampled data are introduced.
Importantly, these perturbations are quickly absorbed within a few update steps, and the loss returns to a low regime.
Compared to the Router, the Refactorer shows a larger variance in early rounds, consistent with the higher difficulty and heterogeneity of generation-style rewriting objectives; however, its fluctuations diminish over time and stabilize near zero, with only sparse spikes on harder conflict-heavy cases.
Overall, the loss trends suggest stable optimization and fast adaptation to round-wise data refreshes.

\section{Implementation Details}
\label{app:impl}

\subsection{System Overview}
\label{app:impl_overview}

Our framework realizes a \textit{self-evolving} training loop that repeatedly alternates between
\textbf{trajectory generation} and \textbf{incremental fine-tuning}.
At inference time, we augment a standard agent with an explicit \textbf{controller} consisting of a
\textbf{Router} (drift diagnosis) and a \textbf{refactorer} (context refactoring).
Given the evolving interaction history $H_t$, the Router decides whether refactoring is needed and selects an operator $o_t \in \mathcal{O}\cup\{\textsc{none}\}$.
If $o_t \neq \textsc{none}$, the refactorer produces a refactored context $\tilde{H}_t$, which is then used to condition the Actor for the next action.
This design separates \textit{context management} from \textit{task solving}, enabling modular training and controlled interventions.

\subsection{Controller I/O Protocol}
\label{app:impl_protocol}

\textbf{History-only diagnosis.} To avoid trivially conditioning on the current query and to encourage robust drift detection,
the Router receives only the accumulated history $H_t$ (and the task description) rather than the current step query.
It outputs a compact decision in a structured form: a binary drift flag and an operator choice.
This protocol makes the Router a \textit{context monitor} rather than a solver.

\textbf{Refactoring as replaceable state.} The refactorer outputs a single refactored context block $\tilde{H}_t$ that is directly replaceable as the ``previous context'' buffer.
We enforce a strict output contract (no extra commentary) to prevent hidden leakage into the Actor prompt and to ensure that refactoring is auditable.

\subsection{Self-Evolving Data Generation}
\label{app:impl_data_gen}

\textbf{Cold-start supervision.} We initialize learning from a small seed set produced by a strong teacher model.
The teacher provides (i) operator decisions for the Router and (ii) refactored contexts for the refactorer.
This stage stabilizes early training by preventing degenerate policies (e.g., always selecting a single operator).

\textbf{Evolution via staged locality.} After cold start, we iteratively improve the controller by generating new trajectories with increasingly local components: in the first evolution iteration, we run a \textit{hybrid} setting where the Router is local (student) while the refactorer remains teacher-guided.
From the second iteration onward, both Router and refactorer operate locally.
This staged schedule mitigates compounding errors: early on, we distill high-quality refactoring behaviors before relying fully on local refactoring.

\textbf{Branch forking for supervision diversity.} When the Router detects drift, we optionally fork an environment state at the refactoring step and continue execution along two branches:
a baseline branch (without refactoring) and a refactored branch.
Forking increases the diversity of corrective examples and allows the system to observe counterfactual outcomes under different context states.
To control computational growth, we bound the maximum refactoring depth per episode and cap the refactoring budget of each trajectory.

\subsection{Training Objective and Optimization}
\label{app:impl_training}

We train Router and refactorer as \textbf{separate} parameter-efficient adapters over a shared base model.
Both modules are optimized with supervised fine-tuning using the collected JSONL pairs.
During evolution, we perform \textit{incremental} adapter updates by initializing from the previous iteration's adapter, which enables continual improvement without retraining from scratch.

\textbf{Early stopping per module.} Since operator selection and context rewriting may converge at different speeds, we track their validation losses independently.
We stop updating a module if its loss fails to improve beyond a minimum margin for several iterations, while allowing the other module to continue evolving.
This prevents overfitting and unnecessary computation once one component saturates.

\subsection{Engineering Choices for Stability and Throughput}
\label{app:impl_engineering}

\paragraph{Efficient local inference with dynamic adapters.}
For fully local evolution, we use an offline inference backend that supports dynamic LoRA switching.
This enables updating Router/refactorer adapters between iterations without reloading the base model, substantially reducing iteration overhead.

\subsection{Reproducibility Details}
\label{app:impl_repro}

\paragraph{Configuration.} All hyperparameters and system choices are specified in a single YAML configuration. We export the resolved configuration alongside training/validation logs for each run.

\paragraph{Data and checkpointing.}
For each phase/iteration, we save the Router and refactorer datasets separately.
We also persist a lightweight checkpoint containing: current iteration, adapter paths, dataset paths, sample counts, and a record of used tasks for de-duplication.
This allows resuming evolution without repeating data collection and helps prevent accidental train-eval leakage.


\clearpage
\onecolumn
\section{Prompts}
\label{app:prompts}

We list the prompt templates used by the search agent, router, and refactorer modules. Please refer to our code base for more details.

\textbf{Placeholders.} We use the following placeholders in our prompt templates:
\texttt{\{task\_description\}} (the task goal or user query),
\texttt{\{history\}} (the raw interaction history between the agent and the environment/user),
\texttt{\{previous\_context\}} (the previously refactored context; set to \texttt{None} if empty),
\texttt{\{step\_count\}} (the number of steps taken so far in the search loop),
and \texttt{\{memory\_context\}} (a structured search memory that concatenates prior \texttt{<search>...</search>} queries and the corresponding \texttt{<information>...</information>} results).
All outputs follow strict structural contracts to enable deterministic parsing.

\subsection{Search Prompts}
We standardize the search-augmented reasoning procedure with a unified \textsc{Search-Agent} prompt. At each step, the model must first write its internal reasoning enclosed in \texttt{<think>...</think>}, and then emit exactly one action: either (i) issue a web query using \texttt{<search>...</search>} when external evidence is needed, or (ii) return the final answer using \texttt{<answer>...</answer>} without additional explanations. For multi-step interactions, we additionally provide the accumulated step count and a structured memory context, where previous queries and retrieved evidence are explicitly tagged with \texttt{<search>} and \texttt{<information>} to support traceable evidence aggregation and prevent mixing retrieval with answering in the same step.

\begin{figure*}[h]
\centering
\resizebox{0.95\textwidth}{!}{
\begin{tcolorbox}[
  colback=gray!5!white,
  colframe=black!75!black,
  title=Search Prompt Template,
  boxrule=0.3mm,
  width=\textwidth,
  arc=3mm,
  auto outer arc=true
]
You are an expert agent tasked with answering the given question step-by-step.

Your question: \{task\_description\}

(Optional) Prior to this step, you have already taken \{step\_count\} step(s). Below is the interaction history, where \texttt{<search>...</search>} wraps your past search queries and \texttt{<information>...</information>} wraps the corresponding results returned by the external search engine:
\{memory\_context\}

Now it's your turn to respond for the current step.

You should first conduct a reasoning process. This process MUST be enclosed within \texttt{<think>} \texttt{</think>} tags.
After completing your reasoning, choose \textbf{only one} of the following actions (do not perform both):
(1) If you lack external knowledge, call a search engine using: \texttt{<search> your query </search>}.
(2) If you have enough knowledge to answer confidently, provide the final answer using: \texttt{<answer> ... </answer>} (no additional explanation).
\end{tcolorbox}}
\end{figure*}

\subsection{Router Prompt}
\label{app:router_prompt}
\begin{figure*}[h]
\centering
\resizebox{0.95\textwidth}{!}{
\begin{tcolorbox}[
  colback=gray!5!white,
  colframe=black!75!black,
  title=Router System Prompt,
  boxrule=0.3mm,
  width=\textwidth,
  arc=3mm,
  auto outer arc=true
]
You are a Context Monitor for multi-turn dialogue LLMs.

Your job is to analyze the accumulated history and detect context drift, i.e., cases where the history becomes noisy, misleading, inconsistent, or excessively long and may harm the LLM's next decision.

Important constraint:
You will receive only the history context, not the current query. Decide whether refactoring would help with the next action.

Available operators:
1) state\_abstract: compress verbose history into a high-level state snapshot. Use when the key state is buried in details.
2) noise\_filter: remove irrelevant or redundant content. Use when the history contains off-topic or repeated text.
3) fact\_rectify: correct contradictions or factual errors. Use when the history conflicts with the current state.
4) path\_prune: remove repeated failures or loops. Use when the history shows circular attempts.
5) cognitive\_boosting: inject a short guiding thought to refocus. Use when the LLM is confused about the goal or next step.
6) attention\_anchor: move or copy critical constraints to the end for recall. Use when important requirements are being overlooked.
7) none: no refactoring needed. Use when the history is clean and focused.

Output format (strict):
Return only a valid JSON object with the following fields:
\{"analysis": "<brief 1-2 sentence explanation>", "drift\_detected": <true or false>, "selected\_operator": "<operator\_name>"\}

Rules:
- If drift\_detected is false, selected\_operator must be "none".
- If drift\_detected is true, selected\_operator must be one of the six active operators.
- Be conservative: only flag drift when it is likely to impact reasoning.
- Output only the JSON object, nothing else.
\end{tcolorbox}
}
\caption{Router system prompt. Markdown markers are removed; the output contract is strictly JSON.}
\label{prompt:router_system}
\end{figure*}

\begin{figure*}[h]
\centering
\resizebox{0.95\textwidth}{!}{
\begin{tcolorbox}[
  colback=gray!5!white,
  colframe=black!75!black,
  title=Router User Message Template,
  boxrule=0.3mm,
  width=\textwidth,
  arc=3mm,
  auto outer arc=true
]
Task description:
\textcolor{brown}{\{task\_description\}}

History context (analyze for drift):
\textcolor{brown}{\{history\}}

Instruction:
Analyze the history and output your assessment as a JSON object.
\end{tcolorbox}
}
\caption{Router input template.}
\label{prompt:router_user}
\end{figure*}

\subsection{Refactorer Prompts}
\label{app:reconstructor_prompts}

\begin{figure*}[h]
\centering
\resizebox{0.95\textwidth}{!}{
\begin{tcolorbox}[
  colback=gray!5!white,
  colframe=black!75!black,
  title=Refactorer Shared System Prompt,
  boxrule=0.3mm,
  width=\textwidth,
  arc=3mm,
  auto outer arc=true
]
You are a Context Refactoring Engine for multi-turn dialogue LLMs.

Your job is to transform the provided history to improve the LLM's next decision by applying a specified transformation operator.

Core principles:
1) Preserve critical information needed for task completion.
2) Maintain coherence: the refactored context must be logically consistent and self-contained.
3) Enable progress: the result should support better decisions going forward.
4) Be conservative: when uncertain, preserve rather than remove.

Output rules:
1) Output only the refactored context within <summary> </summary> tags.
2) The refactored context must be directly usable as the new history.
3) Do not include any explanation or meta-commentary.
4) Do not add information that was not present in the original context.
5) Aim for meaningful compression while keeping critical information.
\end{tcolorbox}
}
\caption{Refactorer shared header used across all operators.}
\label{prompt:recon_header}
\end{figure*}

\textbf{Operator templates.} Given a Router-selected operator $o$, we instantiate an operator-specific user prompt by filling the placeholders
\texttt{\{task\_description\}}, \texttt{\{history\}}, and \texttt{\{previous\_context\}}.

\begin{figure*}[h]
\centering
\resizebox{0.95\textwidth}{!}{
\begin{tcolorbox}[
  colback=gray!5!white,
  colframe=black!75!black,
  title=Operator Prompt: state\_abstract,
  boxrule=0.3mm,
  width=\textwidth,
  arc=3mm,
  auto outer arc=true
]
Role:
You are a precise text processing engine specialized in state abstraction.

Operator name:
State Abstraction

Objective:
Compress the interaction history into a concise state snapshot.

Transformation logic:
1) Identify the net results of actions in history.
2) Remove intermediate steps that no longer matter.
3) Replace detailed sequences with a compact description of the current physical/logical state.
4) Keep the most recent observation and any critical discoveries.

Key principles:
- Focus on what has been achieved, not how it was achieved.
- Track inventory changes.
- Track environment state changes.
- Preserve discovered constraints and rules.

Current task:
\textcolor{brown}{\{task\_description\}}

Previous refactored context:
\textcolor{brown}{\{previous\_context\}}

Input context (raw interaction history):
\textcolor{brown}{\{history\}}

Output format:
Return only:
<summary>
... state snapshot ...
</summary>
\end{tcolorbox}
}
\caption{Operator prompt for \textsc{state\_abstract}.}
\label{prompt:op_state_abstract}
\end{figure*}

\begin{figure*}[h]
\centering
\resizebox{0.95\textwidth}{!}{
\begin{tcolorbox}[
  colback=gray!5!white,
  colframe=black!75!black,
  title=Operator Prompt: noise\_filter,
  boxrule=0.3mm,
  width=\textwidth,
  arc=3mm,
  auto outer arc=true
]
Role:
You are a precise text processing engine specialized in noise filtration.

Operator name:
Noise Filtration

Objective:
Remove irrelevant noise while preserving useful information.

Transformation logic:
1) Identify segments orthogonal to the task.
2) Delete such segments entirely.
3) Ensure the remaining text is coherent and temporally consistent.

Remove:
- Repeated identical observations.
- Verbose descriptions that add no information.
- Failed actions that provide no new constraints.
- Navigation steps that do not change state.

Keep:
- State-changing actions and outcomes.
- New discoveries.
- Constraint-bearing error messages.
- The most recent observation.

Current task:
\textcolor{brown}{\{task\_description\}}

Previous refactored context:
\textcolor{brown}{\{previous\_context\}}

Input context (noisy history):
\textcolor{brown}{\{history\}}

Output format:
Return only:
<summary>
... filtered context ...
</summary>
\end{tcolorbox}
}
\caption{Operator prompt for \textsc{noise\_filter}.}
\label{prompt:op_noise_filter}
\end{figure*}

\begin{figure*}[h]
\centering
\resizebox{0.95\textwidth}{!}{
\begin{tcolorbox}[
  colback=gray!5!white,
  colframe=black!75!black,
  title=Operator Prompt: fact\_rectify,
  boxrule=0.3mm,
  width=\textwidth,
  arc=3mm,
  auto outer arc=true
]
Role:
You are a precise text processing engine specialized in fact rectification.

Operator name:
Fact Rectification

Objective:
Identify and correct inconsistencies or contradictions in the interaction history while preserving correct content.

Trusted signals:
- The current task description defines the goal.
- The most recent observation reflects the true current state.
- Successful actions are factual evidence; failed actions reveal constraints.

Transformation logic:
1) Locate statements in the history that conflict with the current observed state.
2) Detect logical inconsistencies.
3) Edit only the minimal conflicting spans to match trusted signals.
4) Preserve all correct parts of the history unchanged.
5) Do not invent new facts that are not supported by the task or observations.

Common issues to fix:
- Incorrect inventory tracking.
- Wrong location assumptions.
- Misremembered action outcomes.
- Contradictory state descriptions.

Current task:
\textcolor{brown}{\{task\_description\}}

Previous refactored context:
\textcolor{brown}{\{previous\_context\}}

Input context:
\textcolor{brown}{\{history\}}

Output format:
Return only:
<summary>
... rectified context ...
</summary>
\end{tcolorbox}
}
\caption{Operator prompt for \textsc{fact\_rectify}.}
\label{prompt:op_fact_rectify}
\end{figure*}

\begin{figure*}[h]
\centering
\resizebox{0.95\textwidth}{!}{
\begin{tcolorbox}[
  colback=gray!5!white,
  colframe=black!75!black,
  title=Operator Prompt: path\_prune,
  boxrule=0.3mm,
  width=\textwidth,
  arc=3mm,
  auto outer arc=true
]
Role:
You are a precise text processing engine specialized in path pruning.

Operator name:
Path Pruning

Objective:
Truncate the history to remove failed branches and repetitive loops, while preserving any useful discoveries.

Transformation logic:
1) Identify where the interaction begins to loop, stall, or repeatedly fail.
2) Recognize loop patterns such as:
   - trying the same action multiple times with the same failure,
   - repeatedly searching the same locations without new findings,
   - back-and-forth navigation returning to an unchanged state.
3) Delete the looping or dead-end portion.
4) Preserve any new information discovered, even within the failed branch.
5) End the context at a clean decision point so the LLM can attempt a new plan.

Pruning criteria:
- Remove sequences of three or more similar failed actions.
- Remove back-and-forth navigation that returns to the same state.
- Remove repeated searches of empty containers or rooms.
- Keep: discoveries, constraints, and the latest valid state summary.

Current task:
\textcolor{brown}{\{task\_description\}}

Previous refactored context:
\textcolor{brown}{\{previous\_context\}}

Input context (history with potential loops/failures):
\textcolor{brown}{\{history\}}

Output format:
Return only:
<summary>
... pruned context ending at a clean decision point ...
</summary>
\end{tcolorbox}
}
\caption{Operator prompt for \textsc{path\_prune}.}
\label{prompt:op_path_prune}
\end{figure*}

\begin{figure*}[h]
\centering
\resizebox{0.95\textwidth}{!}{
\begin{tcolorbox}[
  colback=gray!5!white,
  colframe=black!75!black,
  title=Operator Prompt: cognitive\_boosting,
  boxrule=0.3mm,
  width=\textwidth,
  arc=3mm,
  auto outer arc=true
]
Role:
You are a precise text processing engine specialized in cognitive reinforcement.

Operator name:
Cognitive Reinforcement

Objective:
Insert a short guiding directive to refocus the LLM and improve the next decision, without changing factual content.

Transformation logic:
1) Identify where the LLM becomes confused, inefficient, or stuck in the history.
2) Determine an actionable next sub-goal based on the task and the current state.
3) Insert a directive formatted exactly as:
   [Thought]: ...
4) The directive must be specific and actionable (what to do next), but must not add unsupported facts.

Reinforcement strategies:
- If stuck: recommend unexplored locations or a different interaction strategy.
- If confused: restate the immediate sub-goal clearly.
- If inefficient: suggest a more direct plan.
- If close to completion: highlight the remaining required steps.

Placement rule:
Insert the [Thought] directive near the end of the refactored context, right before the most recent situation description, so it is salient for the next action.

Current task:
\textcolor{brown}{\{task\_description\}}

Previous refactored context:
\textcolor{brown}{\{previous\_context\}}

Input context:
\textcolor{brown}{\{history\}}

Output format:
Return only:
<summary>
... refactored context ...

[Thought]: ... (one or two sentences, actionable)
</summary>
\end{tcolorbox}
}
\caption{Operator prompt for \textsc{cognitive\_boosting}.}
\label{prompt:op_cognitive_reinforce}
\end{figure*}

\begin{figure*}[h]
\centering
\resizebox{0.95\textwidth}{!}{
\begin{tcolorbox}[
  colback=gray!5!white,
  colframe=black!75!black,
  title=Operator Prompt: attention\_anchor,
  boxrule=0.3mm,
  width=\textwidth,
  arc=3mm,
  auto outer arc=true
]
Role:
You are a precise text processing engine specialized in attention anchoring.

Operator name:
Attention Anchoring

Objective:
Move or copy critical information to the end of the context so it remains in the active attention region for the next action.

Transformation logic:
1) Identify critical information mentioned earlier but essential for completing the task.
2) Typical critical information includes: task objective, constraints, inventory, discovered key locations/items, partial progress, and what has already been searched.
3) Append a final section named [KEY INFO] at the very end of the context.
4) Ensure the [KEY INFO] section is the last content before the model generates the next action.
5) Do not add new facts; only restate or re-organize information that exists in the input.

What to anchor:
- The main task objective.
- Current inventory.
- Locations already searched.
- Discovered constraints or rules.
- Partial progress and remaining steps.

Current task:
\textcolor{brown}{\{task\_description\}}

Previous refactored context:
\textcolor{brown}{\{previous\_context\}}

Input context (history with potentially forgotten information):
\textcolor{brown}{\{history\}}

Output format:
Return only:
<summary>
... refactored context ...

[KEY INFO]:
- Task: ...
- Current inventory: ...
- Searched locations: ...
- Constraints: ...
- Next logical step: ...
</summary>
\end{tcolorbox}
}
\caption{Operator prompt for \textsc{attention\_anchor}.}
\label{prompt:op_attention_anchor}
\end{figure*}

We enforce strict output contracts for reliable deployment. 
For the Router, we parse the output as JSON and validate all required fields, in particular 
\texttt{drift\_detected} and \texttt{selected\_operator} against the predefined operator set. 
If parsing or validation fails, we apply a conservative fallback by setting 
\texttt{drift\_detected=false} and \texttt{selected\_operator="none"}, and pass the raw history forward. 
For the Refactorer, we extract the content within \texttt{<summary>...</summary>} tags; 
if tags are missing or the extracted summary is empty, we fall back to using the raw model output as the refactored context. 
These safeguards prevent error propagation and ensure the pipeline never silently proceeds with an ill-formed refactoring.

\end{document}